%% file: main.tex
\documentclass[10pt,twocolumn,letterpaper]{article}

\usepackage[pagenumbers]{iccv} 

\usepackage{graphicx}
\usepackage{float}
\usepackage{tabularray}


\definecolor{iccvblue}{rgb}{0.21,0.49,0.74}
\usepackage[pagebackref,breaklinks,colorlinks,allcolors=iccvblue]{hyperref}

\def\paperID{12899} 
\def\confName{ICCV}
\def\confYear{2025}

\title{LiM-Loc: Visual Localization with Dense and Accurate 3D Reference Maps Directly Corresponding 2D Keypoints to 3D LiDAR Point Clouds}

\author{Masahiko Tsuji \qquad \quad Hitoshi Niigaki \qquad \quad Ryuichi Tanida\\
NTT Corporation, Japan\\
}

\begin{document}

\twocolumn[{%
\renewcommand\twocolumn[1][]{#1}%
\maketitle
\begin{center}
    \centering
    \captionsetup{type=figure}
    \includegraphics[width=1.0\textwidth]{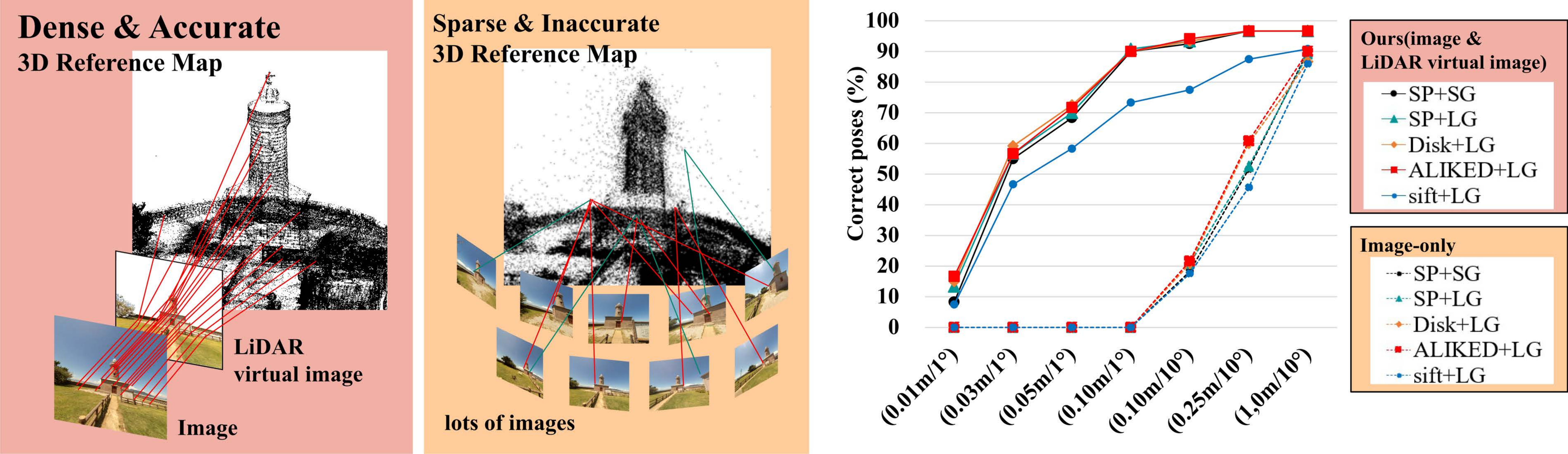}
    \captionof{figure}{LiM-Loc directly assigns 2D keypoints to 3D LiDAR point clouds to generate a dense and accurate 3D reference map. This method collaborates with a variety of state-of-the-art local features to 3D reconstruct almost every keypoint. LiM-Loc estimates with an error of less than a few centimeters, which is difficult with image-only methods.}
\end{center}%
}]

\input{sec/0_abstract}

\input{sec/1_intro}
\input{sec/2_Related}

\input{sec/3_Method}
\input{sec/4_Evaluation}
\input{sec/5_Conclusion}
{
    \small
    \bibliographystyle{ieeenat_fullname}
    \bibliography{main}
}

\end{document}

%% file: sec/0_abstract.tex
\begin{abstract}
Visual localization is to estimate the 6-DOF camera pose of a query image in a 3D reference map. We extract keypoints from the reference image and generate a 3D reference map with 3D reconstruction of the keypoints in advance. We emphasize that the more keypoints in the 3D reference map and the smaller the error of the 3D positions of the keypoints, the higher the accuracy of the camera pose estimation. However, previous image-only methods require a huge number of images,  and it is difficult to 3D-reconstruct keypoints without error due to inevitable mismatches and failures in feature matching. As a result, the 3D reference map is sparse and inaccurate. In contrast, accurate 3D reference maps can be generated by combining images and 3D sensors. Recently, 3D-LiDAR has been widely used around the world. LiDAR, which measures a large space with high density, has become inexpensive. In addition, accurately calibrated cameras are also widely used, so images that record the external parameters of the camera without errors can be easily obtained. In this paper, we propose a method to directly assign 3D LiDAR point clouds to keypoints to generate dense and accurate 3D reference maps. The proposed method avoids feature matching and achieves accurate 3D reconstruction for almost all keypoints. To estimate camera pose over a wide area, we use the wide-area LiDAR point cloud to remove points that are not visible to the camera and reduce 2D-3D correspondence errors. Using indoor and outdoor datasets, we apply the proposed method to several state-of-the-art local features and confirm that it improves the accuracy of camera pose estimation.
\end{abstract}

%% file: sec/1_intro.tex
\section{Introduction}
\label{sec:intro}
Visual Localization aims at estimating the 6-DoF camera pose(i.e. position and orientation) in a 3D reference map\cite{01,02,03,04,05,06,07,40}. It can estimate the camera pose even in places where GPS and GNSS are difficult, such as indoor environments where satellite signals are difficult to reach, or in high-rise buildings where satellite signals are reflected in multiple stages, so it is used for various applications such as autonomous driving, robotics, and augmented reality. In general, camera pose estimation uses a 3D reference map to obtain the 3D positions of keypoints in the query image and then estimates the camera pose using the Perspective-n-Point (PnP) algorithm\cite{08,09}. The 3D reference maps that provide the input for PnP are generated by Structure-from-Motion (SfM)\cite{08,10}. In SfM, common keypoints between multiple reference images are extracted by feature matching, and then the keypoints are 3D reconstructed. The reconstructed keypoints are given 3D positions and stored as a 3D reference map. The 3D reference map provides the input values for PnP\cite{08,09}, which calculates the camera pose, and therefore has a significant impact on the estimation accuracy. However, in SfM, feature matching is affected by insufficient texture or similar texture, leading to incorrect feature matching or failure to match. It is difficult to reconstruct keypoints in 3D without error, and as a result, the 3D reference map is sparse and inaccurate. Increasing the number of reference images will make the 3D reference map generated by SfM denser and more accurate, but it will also increase processing time and have limited quality.

In contrast, combining images and 3D sensors can generate dense and accurate 3D reference maps. Recently, with the increasing demand for autonomous driving, robotics, and AR, 3D-LiDAR has become widely used in the world. High-end iPhone models are equipped with LiDAR, and LiDAR is becoming more common. For example, NavVis-VLX and BLK2GO are portable LiDAR that measured while walking. They can easily measure large indoor and outdoor areas. In addition, they are equipped with accurately calibrated cameras, so images that record camera external parameters with low error can be easily obtained. MMS is a LiDAR that targets a larger space. MMS is a car equipped with LiDAR and a camera and can acquire city-scale LiDAR point clouds and images while driving. These LiDAR measure time-series LiDAR point clouds and sequentially register them, reducing measurement errors and measuring more accurate and denser point clouds. This LiDAR registration also reduces measurement errors in the external parameters of the camera\cite{13,14}. As products with accurate calibration of LiDAR and cameras become more widespread, sensor fusion has become easier.

We emphasize that the more keypoints in the 3D reference map and the smaller the error of the 3D positions of the keypoints, the higher the accuracy of the camera pose estimation. In this paper, we propose a method to directly assign 3D LiDAR points to keypoints to generate a dense and accurate 3D reference map. The proposed method avoids feature matching and almost all keypoints are accurately reconstructed in 3D. The proposed method performs HPR (Hidden Point Removal) with spherical compression on the wide-area LiDAR point cloud to reduce 2D-3D correspondence errors. The 3D LiDAR point cloud is projected onto a 2D LiDAR virtual image. The image and the LiDAR virtual image are perfectly overlapped in the camera screen coordinates, and keypoints at the same coordinate positions are directly matched to the LiDAR point cloud in 2D-3D (Figure 1). The proposed method does not rely on feature matching for 2D-3D correspondence and does not rely on specific local features. That is, the proposed method can be applied to various state-of-the-art local features. To prove that our method improves the accuracy of camera pose estimation, we evaluate it on indoor and outdoor datasets. We apply our method to several state-of-the-art local features and show that it improves the accuracy of estimation.\\
The contributions of this paper are as follows:
\begin{enumerate}
    \item We propose LiM-Loc, which combines LiDAR point clouds and images to generate a dense and accurate 3D reference map. By avoiding feature matching and directly 2D-3D matching keypoints to LiDAR point clouds, almost all keypoints can be accurately assigned to LiDAR point clouds.
    \item  Supports registered LiDAR point clouds, which are more accurate and dense. Registered LiDAR point clouds with time-series LiDAR point clouds suffer from occlusions. We use spherical shell compression to remove hidden points from registered LiDAR point clouds, resolving occlusions and reducing misalignment between keypoints and hidden LiDAR points.
    \item  To prove that a dense and accurate 3D reference map improves the accuracy of camera pose estimation, we evaluate our method using two different datasets, indoor and outdoor. We apply our method to several state-of-the-art local features and show improved performance.
\end{enumerate}

%% file: sec/2_Related.tex
\section{Related Works}
\label{sec:Related Works}
\begin{figure*}[t]
  \centering
  \includegraphics[width=15cm]{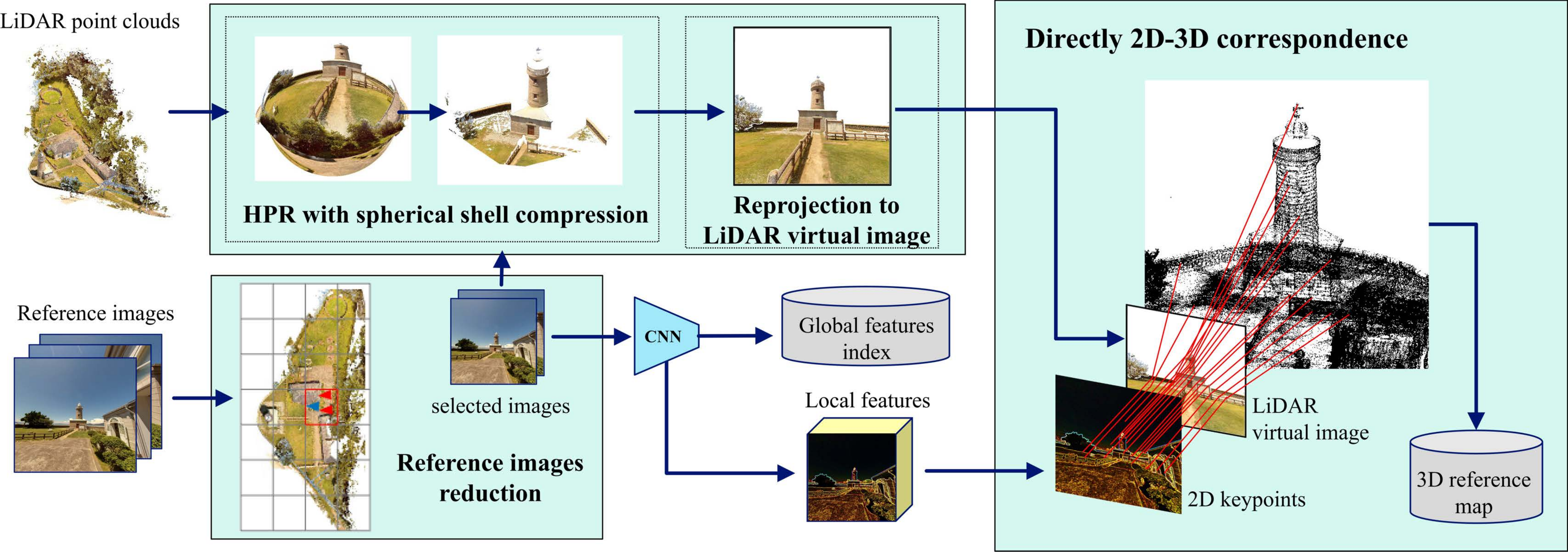}
  \caption{LiM-Loc pipeline consists of (i) reprojecting the LiDAR point cloud onto a realistic LiDAR virtual image, (ii) extracting local image features, and (iii) directly assigning 2D keypoints to the LiDAR virtual image by overlapping them.}
  \label{fig:2}
\end{figure*}

\textbf{Visual localization} is to estimate the camera pose (3D position and orientation) of a query image within a 3D reference map\cite{01,02,03,04,05,06,07,40}. As a preliminary step, we generate a 3D reference map by 3D reconstructing keypoints in the reference image. Generally, in camera pose estimation, the 3D positions of keypoints in the query image are obtained using a 3D reference map, and then the camera pose is estimated using the Perspective-n-Point (PnP) algorithm\cite{08,09}. The 3D reference map that provides the input values for PnP is generated using the Structure-from-Motion (SfM) reconstruction method\cite{08,10,12}. In SfM, common keypoints between multiple reference images are extracted by feature matching, and then the keypoints are 3D reconstructed. The reconstructed keypoints are given 3D positions and stored as a 3D reference map. The 3D reference map has a significant impact on the estimation accuracy, as it provides the input values for PnP\cite{08,09}, which calculates the camera pose. However, in SfM, feature matching is affected by insufficient texture or similar texture, resulting in incorrect feature matching or failure to match. It is difficult to recover keypoints in 3D sparse and inaccurate. Apart from that, SfM requires a huge number of images taken continuously, with a part of the space captured in the previous image also captured without error, and as a result, the 3D reference map is the next image. The hloc\cite{04,05} pipeline extends the task of visual localization to large-scale scenes by adding a global estimation process to extract top-k pairs of reference images from the 3D reference map in image retrieval\cite{01,02,15,16,17}. It has achieved SOTA accuracy in many benchmarks. In this paper, we confirm that the accuracy of camera pose estimation can be improved by applying a dense and accurate 3D reference map to the hloc pipeline.\\
\textbf{Local feature matching} is performed by calculating pairwise descriptor distances between keypoints detected in two images and performing match extraction by mutual nearest neighbors in the distance matrix\cite{19,20,29,31,18}. SuperGlue\cite{05} adopted a graph neural network approach that reflects the information between mutual descriptors and optimal transport instead of mutual nearest neighbors for match extraction and achieved good results in match extraction. However, these methods are not good in texture-less environments where keypoints are difficult to detect. Therefore, semi-dense methods that perform uniform matching on image grids without using a detector have been proposed\cite{33,34,35,36}. They have the advantage of avoiding the problem of detecting keypoints from texture-less images\cite{39}, but if the matching between image grids fails, large image grids will be lost. Recently, dense matching has been proposed to extract feature matching densely across all views\cite{37, 38}. However, while semi-dense and dense approaches have achieved good results for 2D-2D matching, they have not achieved comparable performance to sparse methods \cite{05,22,31} for 3D geometry estimation\cite{21, 37, 38}. Some methods\cite{28, 41,42} directly regress the camera pose from a single query image. But are not competitive in terms of accuracy. Recent studies have improved local feature matching, but these are improvements in the performance of 2D-2D match extraction, and to improve the accuracy of camera pose estimation, it is necessary to improve 2D-3D inliers extraction. Our proposed method avoids feature matching and assigns 3D LiDAR point clouds to keypoints directly, accurately assigning almost all keypoints to 3D LiDAR point clouds. We apply the proposed method to several state-of-the-art local features and confirm that it can improve the accuracy of camera pose estimation.

%% file: sec/3_Method.tex
\section{LiM-Loc}
\label{sec:Proposed Method}

We emphasize that the more keypoints in the 3D reference map and the lower the error of the 3D positions of the keypoints, the higher the accuracy of the camera pose estimation. We propose LiM-Loc, which combines LiDAR point clouds and images to generate a dense and accurate 3D reference map. By avoiding feature matching and directly assigning keypoints to the LiDAR point cloud, almost all keypoints can be 3D reconstructed.

The pipeline of our method is shown in Figure 2. First, we reduce the number of reference images (Reference Image Reduction). Our method can reconstruct almost all keypoints in 3D with a single reference image, so multiple reference images for the same location are not required as in traditional methods. Second, we compress the registered LiDAR point cloud into a spherical shell and remove invisible points (Spherical Shell Compression). The registered LiDAR point cloud has the advantage of being denser and has smaller errors, but it has the disadvantage of occlusion. We solve occlusion and prevent 2D-3D correspondence errors by removing invisible points with Spherical Shell Compression. A 2D LiDAR virtual image is generated by reprojecting the 3D LiDAR point cloud after removing invisible points. The reprojected 2D LiDAR virtual image is perfectly overlapped with the reference image according to the intrinsic and extrinsic parameters of the reference image. Finally, the reference image and the 2D virtual image are associated with keypoints at the same position in the camera screen coordinates (direct 2D-3D correspondence). For camera pose estimation of the query image, we use PnP\cite{08,09}.

\subsection{Reference Image Reduction}
Our method reconstructs almost all keypoints in one reference image in 3D, so ideally, a reference image that comprehensively captures the space without overlapping is sufficient. Reducing the number of reference images can shorten the generation time of the 3D reference map and reduce the data size. However, when measuring, it is difficult to capture images without overlapping in an unfamiliar space. If the reference images are simply reduced after capture, necessary reference images will be lost, resulting in a deterioration of the estimation accuracy of the camera pose. We reduce the reference images while suppressing the deterioration of the estimation accuracy of the camera pose. To suppress the deterioration of the estimation accuracy, it is necessary to suppress the deterioration of the estimation accuracy of both global estimation and local estimation. Therefore, we set an unnecessary metric for each process of global estimation and local estimation and reduce reference images that satisfy both of the two metrics.

\textbf{Global unnecessary metric:} 
In global estimation, the camera pose of the query image is coarsely estimated by image retrieval. In the proposed method, the more similar the camera pose of the reference image is, the less necessary it is. Since a measuring instrument that integrates LiDAR and a camera can obtain an accurate camera pose, a global unnecessary metric is set based on the camera pose of the reference image. First, we divide the 3D reference map into a grid. Next, for reference image $i$ contained in each grid $L_{n}$, we calculate the cosine similarity of the shooting poses $g_{i}$ and $g_{j}$ with other reference images $j$, and this is the global unnecessary metric $C_{ij}$.
\begin{align}
  C_{ij} &= \cos(g_{i},g_{j}), i,j \in L_{n}\\
  &= \frac{ \langle g_{i},g_{j} \rangle }{ \|g_{i},g_{j}\|}
  \label{eq:important}
\end{align}
If the value of (1) exceeds a threshold within the same grid, it is determined to be unnecessary globally.

\textbf{Local unnecessary metric:} In local estimation, locally matching keypoints are extracted between paired images. Images with similar camera pose and many matching keypoints are considered to be of the same location. The local unnecessaryness index $F_{ij}$ is the number of inliers of the local features of paired images $n$ and $m$, and reference images that exceed a threshold are deemed unnecessary. If the function that performs local feature matching and calculates the number of matches is $f(x,y)$, the local index $F_{ij}$ is
\begin{equation}
  F_{ij} = f(i,j), i,j \in L_{n}
\end{equation}

In summary, we divide the space into a grid of equal intervals and calculate the global unnecessary metric $C_{ij}$ and the local unnecessary metric $F_{ij}$ for the images contained in each grid $L_{n}$. We extract paired images for which both metrics are above a threshold. From the extracted paired images, we eliminate the image with fewer local features. This is because the more keypoints that can be input to PnP\cite{08,09}, the easier it is to accurately estimate the camera pose.

\begin{figure}[t]
  \centering
  \begin{subfigure}{1.0\linewidth}
    \includegraphics[width=8cm]{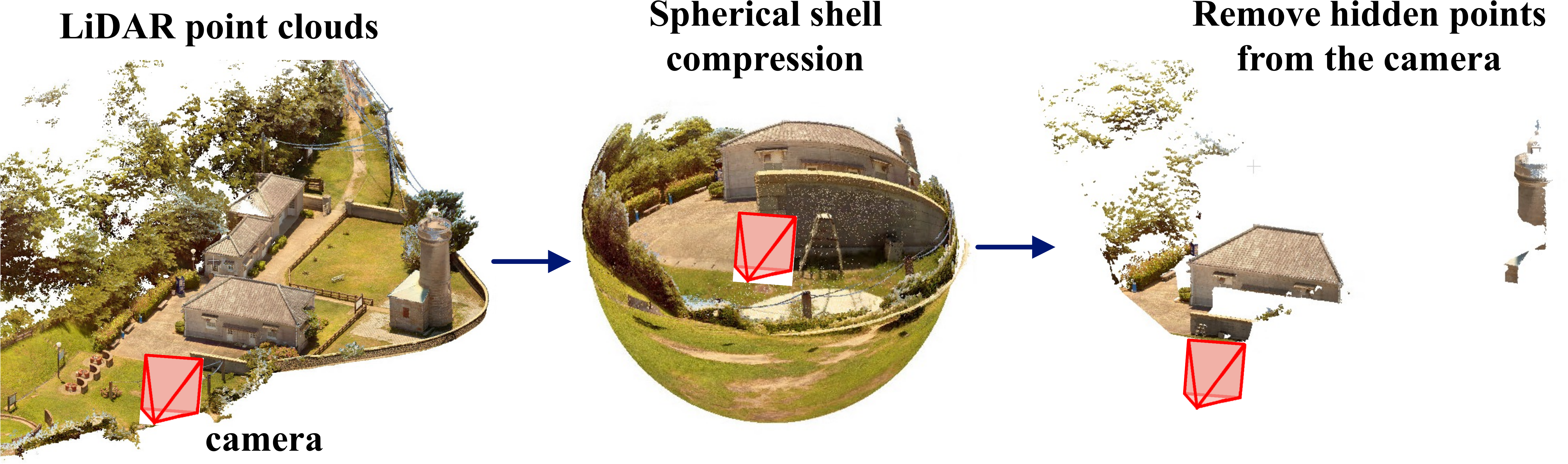}
    \caption{HPR with spherical shell compression.}
    \label{fig3-a}
  \end{subfigure}
  \begin{subfigure}{1.0\linewidth}
    \includegraphics[width=8cm]{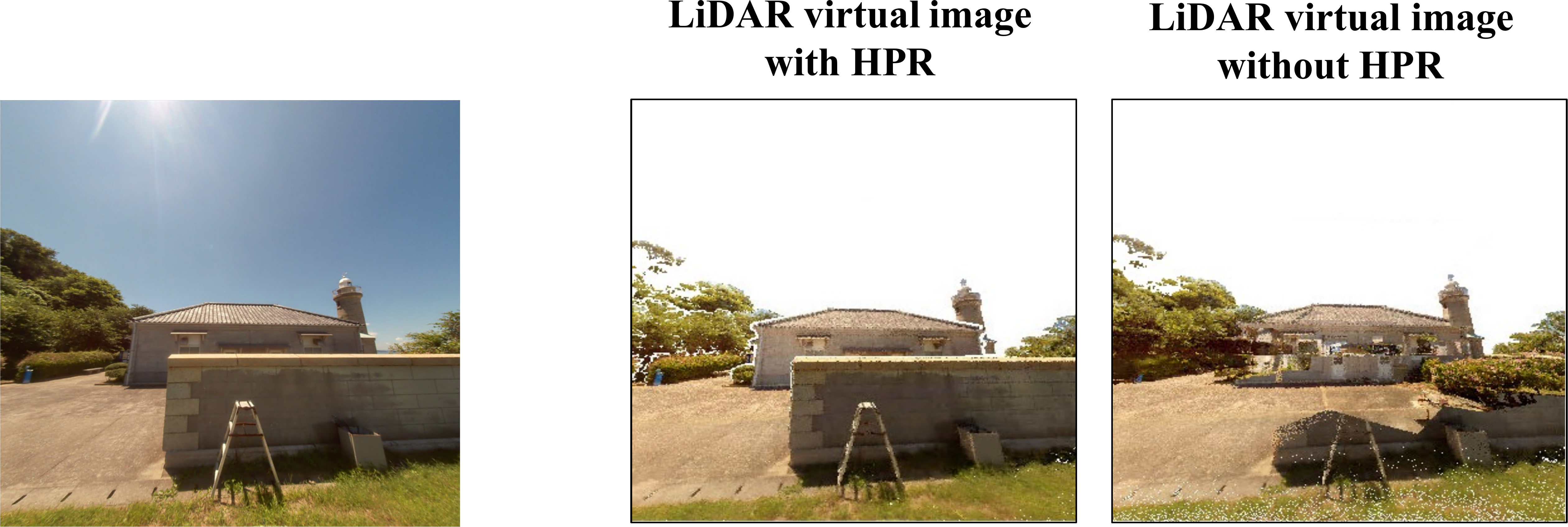}
    \caption{Reprojection to LiDAR Virtual image.}
    \label{fig3-b}
  \end{subfigure}
  \caption{(a) Conventional HPR, which assumes the object-scale, has difficulty in handling spatial-scale point clouds. We compress the spatial-scale point cloud into a spherical shell and convert it into an object-scale point cloud by preserving the visibility from the camera. (b)Without HPR, hidden points appear as noise, which leads to misassignment in 2D-3D correspondence.}
  \label{fig:3}
\end{figure}

\subsection{HPR with spherical shell compression}

\begin{figure*}[t]
  \centering
  \includegraphics[width=17cm]{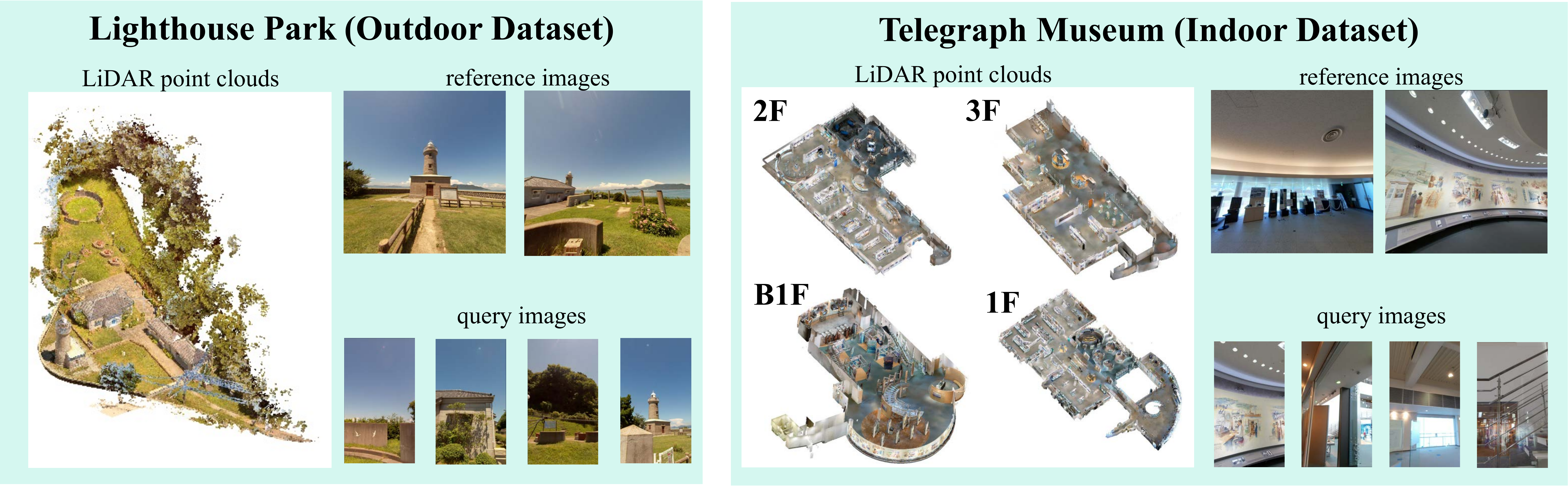}
  \caption{We measured outdoor and indoor datasets using LiDAR while walking. Equipped with a perfectly calibrated camera, we were able to easily capture images with extrinsic parameters.}
  \label{fig:dataset}
\end{figure*}

A registered LiDAR point cloud is a wide-area LiDAR point cloud that represents the entire space, rather than fragmented data measured in one place. Its advantage is a high density of points and low measurement error, which is important when generating dense and accurate 3D reference maps. However, it also has a disadvantage: occlusion. Since we overlay the image and the virtual LiDAR image and directly assign keypoints to the LiDAR point cloud, occlusion will cause invisible points to appear in the virtual LiDAR image Figure 3(b), and keypoints will not match the invisible points.

As a step before generating the virtual LiDAR image, we remove invisible points from the registered LiDAR point cloud. This solves occlusion. A great way to remove invisible points \cite{23} is to flip the point cloud over a virtual sphere and determine the convex hull to determine invisible points. However, this method assumes that the object can be contained within a virtual sphere, so it does not work with the registered point cloud that represents the entire space because the scale is too large. Therefore, we shell centered on the camera position Figure 3(a). By compressing toward the camera center, we preserve the visibility from the camera and shrink the registered LiDAR, and compress the registered LiDAR point cloud into a spherical shell. First, we transform the point cloud into camera-centered coordinates according to the external parameters. Next, the point cloud is compressed into a spherical shell centered on the coordinate origin as
\begin{equation}
  t = s_{max} - s_{min}
  \label{eq:important2}
\end{equation}
where $s_{max}$ ,$s_{min}$ are the distances from the origin to the inner and outer surfaces of the shell. $t$ is represents the thickness of the shell. $P_{n}$ is the 3D position of each point n in the LiDAR point cloud in the camera center coordinates. The distance from the origin is $p^{n}_{norm}$. The distance from the origin to the closest point is $P_{min}$, and the distance from the origin to the farthest point is $P_{max}$. The compression ratio $Comp_{n}$ is
\begin{equation}
  Comp_{n} =\frac{p_{norm}^n - P_{min}}{P_{max} - P_{min}}t + s_{min}
  \label{eq:important3}
\end{equation}
The point cloud $P_{n}$ is transformed into a spherical shell by
\begin{equation}
  P'_{n} = Comp_{n}\frac{p_{n}}{p^n_{norm}}
  \label{eq:important4}
\end{equation}

We apply the spherically compressed point clouds to a previous method\cite{23} to remove hidden points from the wide-scale registered point cloud.

\subsection{Direct 2D-3D correspondence}
The LiDAR point cloud with HPR is projected onto the LiDAR virtual image Figure 2. We make each point of the LiDAR virtual image retain its position in the camera screen coordinates and its original 3D position. Since the reference image and the LiDAR virtual image are unified in the camera screen coordinates, we directly assign the points of the LiDAR virtual image at the positions of the keypoints. Since the point cloud assigned to the keypoints retains its original 3D position, we store it in the 3D reference map as the 3D position of the keypoint.

%% file: sec/4_Evaluation.tex
\section{Experiments}
\label{sec:Experiments}
We confirm that a dense and accurate 3D reference map improves the accuracy of camera pose estimation. We compare the estimation accuracy of several state-of-the-art local features when generating a 3D reference map using our proposed method with that of a conventional method that generates a 3D reference map using only images. 

\subsection{Implementation detail}
Based on the hloc\cite{04,05} pipeline, we replaced the 3D reference map generation part with our proposed method. The pipeline of hloc\cite{04,05} has two parts: an offline process that extracts global features and generates a 3D reference map, and an online process that estimates the camera pose of the query image in real time. We implemented it to generate a 3D reference map in the same format as the 3D reference map output by the offline process of hloc. We did not change the functions of the online process of hloc, but only replaced the 3D reference map. 

\begin{table*}[t]
\begin{tblr}{
  rows={abovesep=-2.0pt,belowsep=-2.0pt},
  columns= {leftsep=1pt,rightsep=1pt},
  width = { 1.0\linewidth },
  hline{1,Z} = { 0.08em }, 
  hline{13} = { 0.08em },
  hline{3} = {1}{-}{},
  hline{3} = {2}{-}{},
  vline{9} = { dotted }, 
  colspec = { X[3]X[1]X[1]X[1]X[1]X[1]X[1]X[1]X[1]X[1]X[1]X[1]X[1]X[1]X[1] }, 
  row{1}= { halign = c, valign = m, font = {\footnotesize} }, 
  row{2-12}= {2-Z}{halign = c, valign = m, font = {\scriptsize} },
  row{13-22}= {2-Z}{halign = c, valign = m, font = {\bfseries\scriptsize} },
  cell{13,18,21,22}{2} = {halign = c, valign = m, font = {\scriptsize} },
  cell{22}{9} = {halign = c, valign = m, font = {\scriptsize} },
  column{8,15}= {halign = c, valign = m, font = {\scriptsize} },
  cell{17}{8} = {halign = c, valign = m, font = {\bfseries\scriptsize} },
  cell{22}{15} = {halign = c, valign = m, font = {\bfseries\scriptsize} },
  column{1}= {3-12}{ halign = l, valign = m, font = {\scriptsize} }, 
  cell{13-22}{1}={ halign = l, valign = m, font = {\bfseries\scriptsize} }, 
  cell{1}{2} = { r = 1, c = 7 }{ halign = c, valign = m }, 
  cell{1}{9} = { r = 1, c = 7 }{ halign = c, valign = m }, 
  cell{1}{1} = { r = 2, c = 1 }{ halign = c, valign = m }, 
}
 Methods & Lighthouse Park &  & &  &  &  &  & Telegraph Museum &  &  &  &  &  &  \\
 &0.01m/1$^{\circ}$ &0.03m/1$^{\circ}$ & 0.05m/1$^{\circ}$ & 0.10m/1$^{\circ}$ & 0.10m/10$^{\circ}$ & 0.25m/10$^{\circ}$ & 1.0m/10$^{\circ}$ & 0.01m/1$^{\circ}$ & 0.03m/1$^{\circ}$ & 0.05m/1$^{\circ}$ & 0.10m/1$^{\circ}$ & 0.10m/10$^{\circ}$ & 0.25m/10$^{\circ}$ & 1.0m/10$^{\circ}$\\
 SP+SG/NV & 0.0 & 0.0 & 0.0 & 0.0 & 18.3 & 51.7 & 90.0 & 0.0 & 0.0 & 0.0 & 0.0 & 18.2 & 51.9 & 85.1\\
 SP+SG/EP & 0.0 & 0.0 & 0.0 & 0.0 & 21.7 & 50.8 & 88.3 & 0.0 & 0.0 & 0.0 & 0.0 & 14.3 & 50.6 & 81.8\\
 SP+LG/NV & 0.0 & 0.0 & 0.0 & 0.0 & 20.8 & 52.5 & 89.2 & 0.0 & 0.0 & 0.0 & 0.0 & 20.8 & 53.9 & 81.8\\
 SP+LG/EP & 0.0 & 0.0 & 0.0 & 0.0 & 16.7 & 49.2 & 88.3 & 0.0 & 0.0 & 0.0 & 0.0 & 18.2 & 51.3 & 79.9\\
 DISK+LG/NV & 0.0 & 0.0 & 0.0 & 0.0 & 20.8 & 60.0 & 86.7 & 0.0 & 0.0 & 0.0 & 0.0 & 14.9 & 43.5 & 75.3\\
 DISK+LG/EP & 0.0 & 0.0 & 0.0 & 0.0 & 15.8 & 61.7 & 87.5 & 0.0 & 0.0 & 0.0 & 0.0 & 13.6 & 39.0 & 75.3\\
 ALIKED+LG/NV & 0.0 & 0.0 & 0.0 & 0.0 & 21.7 & 60.8 & 90.0 & 0.0 & 0.0 & 0.0 & 0.0 & 18.8 & 52.6 & 83.8\\
 ALIKED+LG/EP & 0.0 & 0.0 & 0.0 & 0.0 & 25.0 & 57.5 & 90.8 & 0.0 & 0.0 & 0.0 & 0.0 & 13.6 & 49.4 & 76.0\\
 SIFT+LG/NV & 0.0 & 0.0 & 0.0 & 0.0 & 17.5 & 45.6 & 86.0 & 0.0 & 0.0 & 0.0 & 0.0 & 5.2 & 33.1 & 75.3\\
 SIFT+LG/EP & 0.0 & 0.0 & 0.0 & 0.0 & 14.8 & 41.7 & 83.5 & 0.0 & 0.0 & 0.0 & 0.0 & 6.5 & 31.8 & 68.2\\
 Ours(SP+SG/NV) & 8.3 & 55.0 & 68.3 & 90.0 & 92.5 & 96.7 & 96.7 & 21.4 & 65.6 & 78.6 & 83.8 & 84.4 & 86.4 & 87.0\\
 Ours(SP+SG/EP) & 14.2 & 55.8 & 69.2 & 90.0 & 92.5 & 96.7 & 96.7 & 22.1 & 68.8 & 78.6 & 83.1 & 85.7 & 86.4 & 86.4\\
 Ours(SP+LG/NV) & 13.3 & 56.7 & 70.0 & 90.8 & 93.3 & 96.7 & 96.7 & 21.4 & 66.9 & 78.6 & 83.8 & 85.1 & 85.7 & 85.7\\
 Ours(SP+LG/EP) & 13.3 & 55.8 & 70.8 & 88.3 & 90.8 & 95.8 & 95.8 & 25.3 & 66.9 & 76.0 & 80.5 & 82.5 & 85.7 & 86.4\\
 Ours(DISK+LG/NV) & 15.0 & 59.2 & 72.5 & 90.0 & 93.3 & 96.7 & 96.7 & 16.2 & 51.9 & 64.3 & 72.1 & 74.0 & 81.2 & 83.8\\
 Ours(DISK+LG/EP) & 9.2 & 60.0 & 69.2 & 84.2 & 87.5 & 94.2 & 95.8 & 16.9 & 49.4 & 61.7 & 70.8 & 74.7 & 80.5 & 82.5\\
 Ours(ALIKED+LG/NV) & 16.7 & 56.7 & 71.7 & 90.0 & 94.2 & 96.7 & 96.7 & 11.7 & 61.7 & 72.7 & 81.8 & 83.1 & 83.8 & 85.1\\
 Ours(ALIKED+LG/EP) & 11.7 & 54.2 & 71.7 & 89.2 & 94.2 & 97.5 & 97.5 & 16.2 & 56.5 & 67.5 & 76.6 & 80.5 & 83.8 & 84.4\\
 Ours(SIFT+LG/NV) & 7.5 & 46.7 & 58.3 & 73.3 & 77.5 & 87.5 & 90.8 & 11.0 & 43.5 & 60.4 & 72.7 & 75.3 & 81.2 & 84.4\\
 Ours(SIFT+LG/EP) & 9.2 & 47.5 & 60 & 75.8 & 79.2 & 93.3 & 93.3 & 9.1 & 44.8 & 59.1 & 72.1 & 75.3 & 81.8 & 84.4 
\end{tblr}
\caption{In cases where the same features are used, the results where the estimation accuracy was improved by more than 10\% are in bold(NV:NetVLAD, EP:EigenPlaces). Our method  improves accuracy in cases where the error is smaller than (0.10m/1$^{\circ}$).}
\label{table:accuracy}
\end{table*}

\subsection{Benchmark Datasets}
There are several datasets that contain 3D sensor data\cite{24,25,26,27,40}, but no datasets of images with registered LiDAR point clouds and accurate camera poses. We measured indoors and outdoors with a recent LIDAR, NavVis VLX , and prepared a dataset of registered LiDAR point clouds and images with accurate camera poses. The dataset includes i) registered LiDAR point clouds (.ply), ii) reference images (.jpg), iii) extrinsic parameters of the reference images (.json), iv) internal parameters of the reference image camera (.json), v) query images (.jpg), vi) extrinsic parameters of the query image (.json), and vii) internal parameters of the query image camera (.json). These data are automatically output when measured with NavVis VLX(Figure 5). However, for ii) reference images and v) query images, we reproject them from fisheye images to pinhole images. This is because NavVis VLX outputs fisheye images, which are difficult to apply to SfM. The proposed method can also apply fisheye images, but the evaluation conditions are unified with SfM to use pinhole images. [Lighthouse Park] is the outdoor dataset of a nature park with a small building. The area is approximately 1,550 , and 2,974 reference images and 105 query images are prepared. The indoor dataset [Telegraph Museum] is a museum of information and communication technology. It has four floors from B1F to 3F, and the total area is approximately 5,184 . 3,266 reference images and 154 query images are prepared.

\subsection{Metrics}
We report the percentage of query images that are localized with an error between the estimated camera pose and the ground truth below a threshold. We set seven thresholds: (0.01m, 1.0$^{\circ}$), (0.03m, 1.0$^{\circ}$), (0.05m, 1.0$^{\circ}$), (0.1m, 1.0$^{\circ}$), (0.10m, 10$^{\circ}$), (0.25m, 10$^{\circ}$), and (1.0m, 10$^{\circ}$). The most stringent evaluation method for indoor in the Long-Term Visual Localization Challenge has a minimum error of (0.1m, 1.0$^{\circ}$). We set smaller errors: (0.01m, 1.0$^{\circ}$), (0.03m, 1.0$^{\circ}$), and (0.03m, 1.0$^{\circ}$). The pose error is the average relative rotation error (RRE ($^{\circ}$)).

\begin{table}[h]
  \begin{tblr}{
    rows={abovesep=-2.0pt,belowsep=-2.0pt},
    width = { 1.0\linewidth },
    colspec = { X[1]X[1]X[1] },
    hline{1,Z} = {0.08em }, 
    hline{2} = {0.04em }, 
    row{1,2,3}= { halign = c, valign = m, font = {\footnotesize  } }, 
    row{4}= { halign = c, valign = m, font = {\bfseries\footnotesize } }, 
  }
    & Reference images & Average time [hour] \\
    hloc/COLMAP & 2,974 / 3,266 & 53.3 / 29.9\\
    Ours(No-RIR) & 2,974 / 3,266 & 20.4 / 64.7\\
    Ours & 457 / 1,508 & 3.3 / 26.4
  \end{tblr}
  \caption{The fewer the number of reference images, the faster the average generation time of the 3D reference map. The left side of each cell is outdoors, and the right side is indoors.}
  \label{table-2}
  \end{table}

\subsection{Comparison Condition}
reference map using a smaller number of reference images selected by image reduction and the 3D-LiDAR point cloud. The number of reference images after reduction is x (outdoor dataset) and x (indoor dataset). In contrast, hloc\cite{04,05} gene rates a 3D reference map using all the original 
reference images without using the 3D-LiDAR point cloud. In generating a 3D reference map, the extrinsic and intrinsic parameters of the reference images are important. To make a fair comparison, both the proposed method and 
the conventional method use the extrinsic and intrinsic parameters of the reference images.\\
\textbf{online process:} We did not change the functions of the online process, but only replaced the 3D reference map. The online processing part is no different from hloc\cite{04,05}. but allows the data of the 3D reference map to be replaced.

\subsection{Results}
\textbf{Accuracy:} We evaluated the estimation accuracy of the proposed method. We compared the original method with the proposed method in 10 patterns combining 2 type of global features\cite{01,02} and 5 type of local features/matching\cite{03,06,18,31,05,22}. The results are summarized in Table 1. Regardless of the outdoor and indoor dataset, the proposed method improved the accuracy of the original method in all 10 patterns. The strictest minimum error for indoor environments in the Long-Term Visual Localization Challenge is (0.1 m, 1.0$^{\circ}$). Even with stricter errors (0.01 cm/1.0$^{\circ}$), (0.03 cm/1.0$^{\circ}$), and (0.05 cm/1.0$^{\circ}$), the proposed method can estimate a high rate. In contrast, the difference in estimation accuracy between the proposed method and the conventional method becomes smaller as the error threshold increases. We found that the use of a dense and accurate 3D reference map increases the estimation rate of errors less than (0.1 m/1$^{\circ}$).

\textbf{3D reference map processing time:} We confirmed that the generation time of the 3D reference map is shortened by the RIR (Reference Image Reduction) process. The results are summarized in Table 2. We can see that the generation time is shortened in proportion to the number of reference images reduced by RIR. In previous image-only methods, the more reference images there are, the higher the quality of the 3D reference map (keypoint density and low error). In contrast, our proposed method does not inherently require many reference images. This allows the 3D reference map to be generated in a shorter time.

\begin{table}[t]
  \begin{tblr}{
    rows={abovesep=-2.0pt,belowsep=-2.0pt},
    columns= {leftsep=1pt,rightsep=1pt},
    width = { 1.0\linewidth },
    colspec = { X[1]X[1]X[1]X[1]X[1]X[1]X[1]X[1]},
    hline{1,Z} = {0.08em }, 
    hline{2} = {0.04em }, 
    row{1,2}= { halign = c, valign = m, font = {\scriptsize  } }, 
    row{3,4}= { halign = c, valign = m, font = {\bfseries\scriptsize } },
    cell{3-4}{2} = {halign = c, valign = m, font = {\scriptsize} },
    cell{3}{4} = {halign = c, valign = m, font = {\scriptsize} },
    cell{4}{6,7} = {halign = c, valign = m, font = {\scriptsize} },
  }
    & 0.01m/1$^{\circ}$  & 0.03m/1$^{\circ}$ & 0.05m/1$^{\circ}$ & 0.10m/10$^{\circ}$ & 0.25m/10$^{\circ}$ & 1.0m/10$^{\circ}$  \\
    Ours (SP+SG) & 8.3 / 21.4 & 55.0 / 65.6 & 68.3 / 78.6 & 90.0 / 83.8 & 96.7 / 86.4 & 96.7 / 87.0\\
    No-RIR & 9.2 / 21.4 & 40.0 / 48.7 & 60.0 / \textbf{54.5} & 76.7 / 58.4 & 81.7 / 59.1 & 82.5 / 59.1 \\
    No-HPR & 5.8 / \textbf{5.2} & 38.3 / 30.5 & 58.3 / 40.9 & 78.3 / 46.1 & 87.5 / \textbf{55.2} & 89.2 / \textbf{61.0}
  \end{tblr}
  \caption{This is the estimated accuracy when RIR or HPR is skipped. The left side of each cell is the estimated accuracy outdoors, and the right side is the estimated accuracy indoors. Accuracy degradation of more than 10\% is highlighted in bold.}
  \label{table-3}
\end{table}

\begin{figure}[t]
  \centering
  \begin{subfigure}{1.0\linewidth}
    \includegraphics[width=8cm]{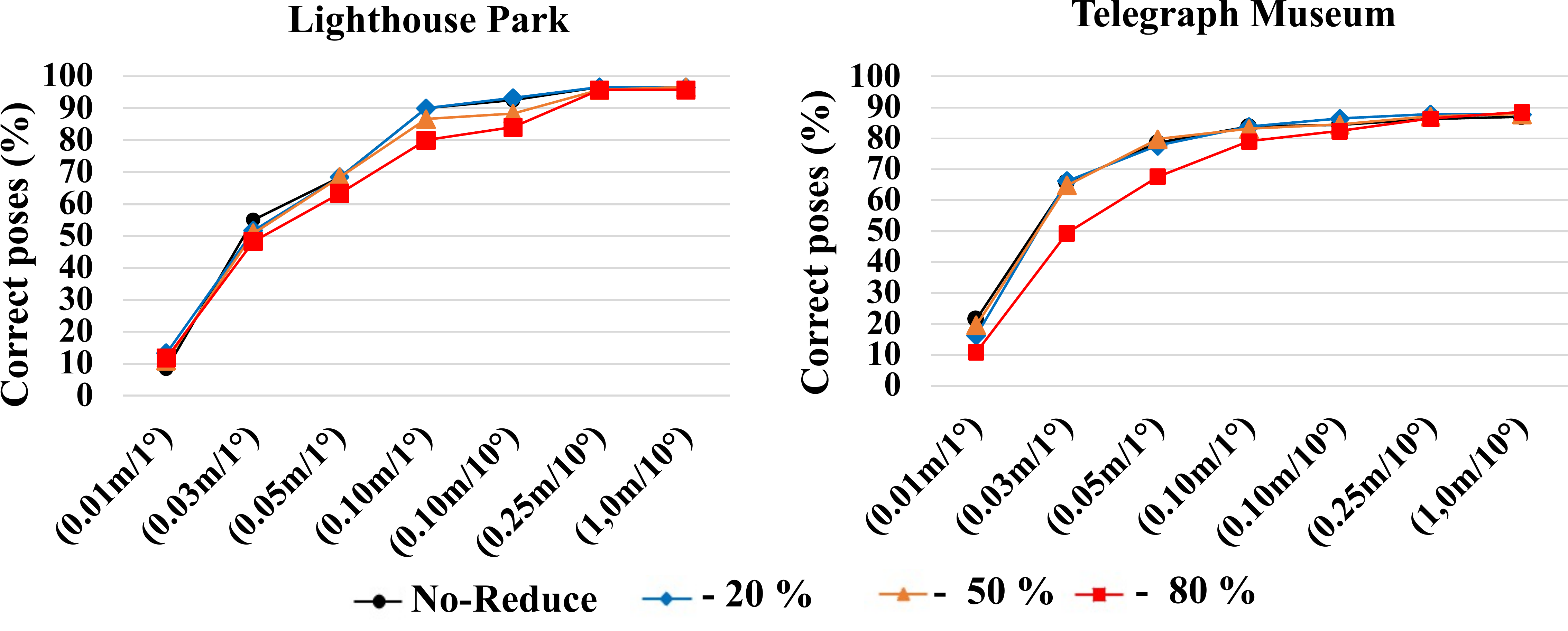}
    \caption{Reduce the number of keypoints}
  \end{subfigure}
  \begin{subfigure}{1.0\linewidth}
    \includegraphics[width=8cm]{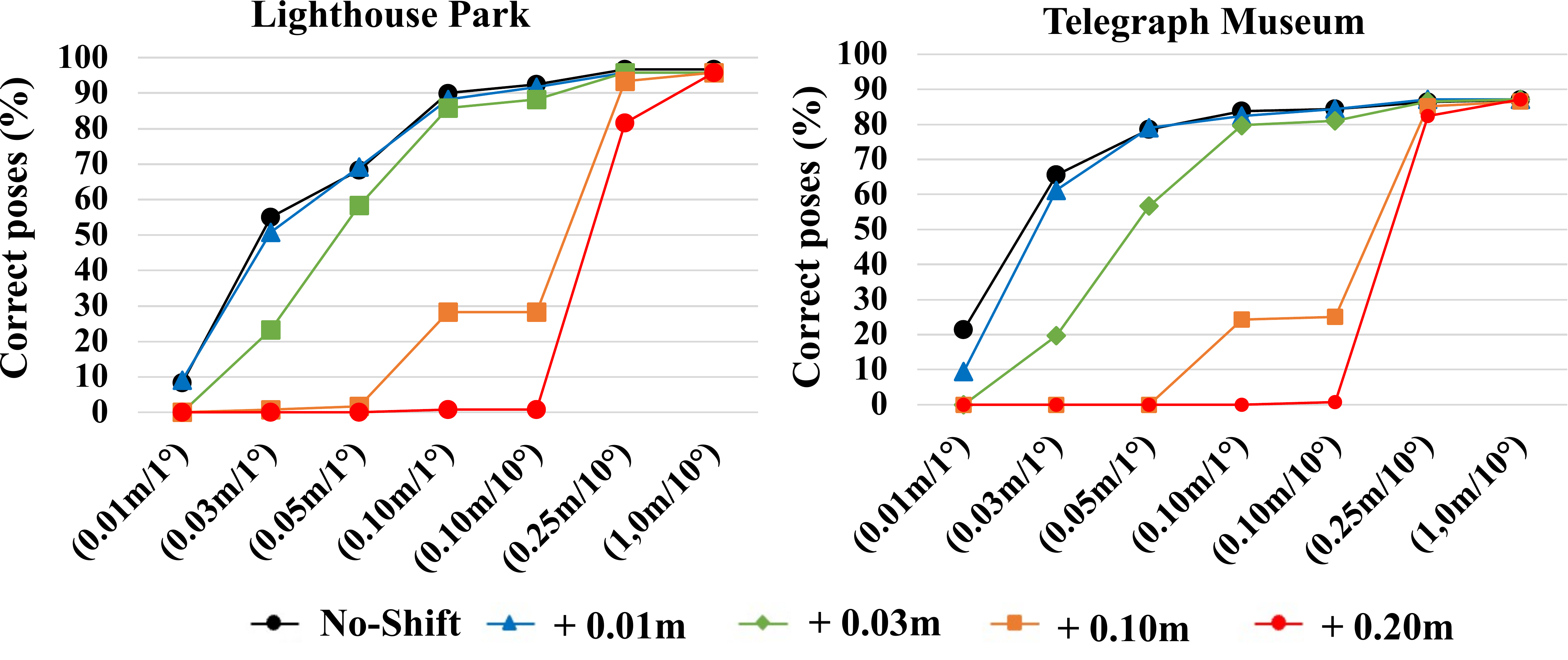}
    \caption{Shift keypoints position}
  \end{subfigure}
  \caption{The impact of (a) reducing the number of keypoints and (b) increasing the 3D position error of keypoints is seen both indoors and outdoors. When the number of keypoints is reduced by 80\% accuracy atarts to degrade. When the 3D position error is larger than 0.10m, estimation below 0.10m become difficult.}
\end{figure}

\subsection{Ablation study}
\textbf{Reference Image Reduction(RIR):} We confirmed whether the proposed reference image reduction can suppress the deterioration of estimation accuracy(Table 3). When the reference images were simply thinned out and reduced in size, the accuracy deteriorated both outdoors and indoors. Indoors, the accuracy deteriorated by about 27.9\% even at the error threshold (1.0 m/10$^{\circ}$). Indoor datasets are often obstructed by obstacles (walls, exhibits, etc.). Compared to outdoors, where there are fewer obstacles blocking the view, more reference images are required indoors. Therefore, there are fewer unnecessary reference images indoors, and simply thinning out reference images increases the possibility of reducing reference images that should not be reduced. The number of reference images required varies depending on the environment, such as the number of obstacles blocking the view, and it is difficult to determine how many to keep. RIR not only has the effect of suppressing deterioration of estimation accuracy, but also automatically keeps the necessary number of images.

\textbf{HPR with spherical compression:} We confirmed whether HPR with spherical shell compression contributes to the estimation accuracy(Table 3). Without HPR, the estimation accuracy deteriorates both outdoors and indoors. Indoors, the estimation accuracy deteriorates more than outdoors. This is because occlusion is more likely to occur in the indoor dataset. The indoor dataset combines four floors, B1F to 3F, into one 3D reference map, so there are Without HPR, there are many invisible points on the LiDAR virtual image. In the vertical direction, LiDAR point clouds from other floors are captured, while in the estimated direction, the back of the walls and exhibits on the same floor are captured. In contrast, in the outdoor area, although there are lighthouses and buildings, there is no occlusion in the vertical direction, and occlusion in the horizontal direction is relatively unlikely to occur. As this result shows, HPR is necessary when dealing with LiDAR point clouds over a wide area. HPR contributes to improving the estimation accuracy.\\

\subsection{Analysis}
What we emphasize is that the more the number of keypoints in the 3D reference map is and the smaller the error in the 3D position of the keypoints is, the higher the accuracy of the camera pose estimation. We show that the number of keypoints and the error in their position affect the estimation accuracy. We check the estimation accuracy when the number of keypoints in the 3D reference map generated by the proposed method is reduced and when the 3D position of the keypoints is shifted. The global features and local features are set to NetVLAD/SP+SG.

\textbf{The number of 3D reconstructed keypoints:} Figure 5(a) shows the estimation accuracy when the number of keypoints in the 3D reference map is reduced to [-20\%, -50\%, -80\%]. To avoid reducing many keypoints from some reference images, we reduced the keypoints equally for each reference image. When the number of keypoints is reduced by 80\%, the accuracy decreases by about 10-16\% below (0.1m/10$^{\circ}$). On the other hand, there is no difference in accuracy above (0.25m/10$^{\circ}$). When the number of keypoints falls below a certain number, the accuracy decreases and it becomes difficult to estimate with an error of less than 0.1m.

\textbf{3D position error:} Figure 5(b) is the estimation accuracy when the 3D position of the keypoint is shifted [0.01m, 0.03m, 0.10m, 0.20m]. To avoid biased shifting in one direction, we shifted 33.3\% of all keypoints in the x direction, 33.3\% in the y direction, and 33.3\% in the z direction. The estimation accuracy does not deteriorate even if the shift is 0.01m. The estimation accuracy deteriorates when the shift is 0.03m, and the accuracy of (0.10m/10$^{\circ}$) is about 1\% when the shift is 20cm. The estimation accuracy when shifted 20cm tends to be similar to that of the conventional method, and it is difficult to estimate less than 0.10m.

So far, we have simulated the degradation of the 3D reference map of the proposed method to confirm the effect of keypoint position and error. Here, we analyze the difference between the 3D reference map of the image-only method and the proposed method. First, the number of 3D reconstructed keypoints is summarized in Table 4. The proposed method has a high ratio of 3D reconstructed keypoints per reference image, about 87-93\%. In contrast, the ratio of the image-only method is low, about 15-38\% outdoors and about 7-22\% indoors. As mentioned above, the image-only method may have a degradation in estimation accuracy due to the small number of keypoints in the 3D reference map.Next, we analyze the 3D position error. Since the proposed method and the conventional method used the same reference image to generate the 3D reference map, Figure.6 summaries the 3D position error of keypoints at the same epixel position on the same reference image and summarized it as a cumulative graph. The percentage of the number of keypoints with an error of less than 0.10 m was about 16\% outdoors and about 62\% indoors. As mentioned above, the degradation of estimation accuracy may occur due to the 3D position error of the keypoints in the 3D reference map.
In summary, the image-only method is likely to be affected by both the number of keypoints and the 3D position error. In particular, the 3D position error of the keypoints is important, and has a large impact on the ratio that can be estimated with an error of less than 0.10 m.

\begin{table}
  \begin{tblr}{
    rows={abovesep=-2.0pt,belowsep=-2.0pt},
    columns= {leftsep=1pt,rightsep=1pt},
    width = { 1.0\linewidth },
    colspec = { X[1]X[1]X[1]X[1]},
    hline{1,Z} = {0.08em }, 
    hline{2} = {2-4}{0.02em }, 
    hline{8} = {0.02em }, 
    hline{3} = {1}{-}{},
    hline{3} = {2}{-}{},
    row{1-7}= { halign = c, valign = m, font = {\scriptsize} }, 
    row{8-12}= { halign = c, valign = m, font = {\bfseries\scriptsize} },
    column{2-3}= { halign = c, valign = m, font = {\scriptsize} },
    cell{1}{2} = { r = 1, c = 3 }{ halign = c, valign = m }, 
    cell{1}{1} = { r = 2, c = 1 }{ halign = c, valign = m }, 
    column{1}= {3-7}{ halign = l, valign = m, font = {\scriptsize} }, 
    cell{8-12}{1}={ halign = l, valign = m, font = {\bfseries\scriptsize} }, 
  }
    & Average per reference image& & \\
    & 2D keypoints & 2D $\to$ 3D keypoints & 2D $\to$ 3D ratio[\%] \\
    Sp+SG & 2,598 / 2,171 & 819 / 414 & 31.5 / 19.1 \\
    Sp+LG & 2,598 / 2,175 & 828 / 428 & 31.9 / 19.7 \\
    Disk+LG & 18,639/19,511 & 6,227 / 3,333 & 33.4 / 17.1 \\
    Aliked+LG & 3,722 / 1,834 & 1,432 / 407 & 38.5 / 22.2 \\
    Sift+LG & 4,536 / 1,635 & 962 / 180 & 21.2 / 11.0 \\
    Ours(SP+SG) & 2,924 / 2,285 & 2,668 / 1,992 & 91.3 / 87.2 \\
    Ours(SP+LG) & 2,924 / 2,285 & 2,668 /1,992 & 91.3 / 87.2 \\
    Ours(Disk+LG) & 19,852/20,352 & 18,344 /18,236 & 92.4 / 89.6 \\
    Ours(Aliked+LG) & 3,784 / 1,887 & 3,535 / 1,701 & 93.4 / 90.1 \\
    Ours(Sift+LG) & 4,727 / 1,669 & 4,363 / 1,484 & 92.3 / 88.9
  \end{tblr}
  \caption{The number of 2D keypoints that could be reconstructed in 3D for each reference image. The left side of each cell is outdoors and the right side is indoors. Our method has a high ratio of 2D$\to$3D about 87-93\%.}
  \label{table-4}
\end{table}

\begin{figure}[t]
  \centering
  \includegraphics[width=8cm]{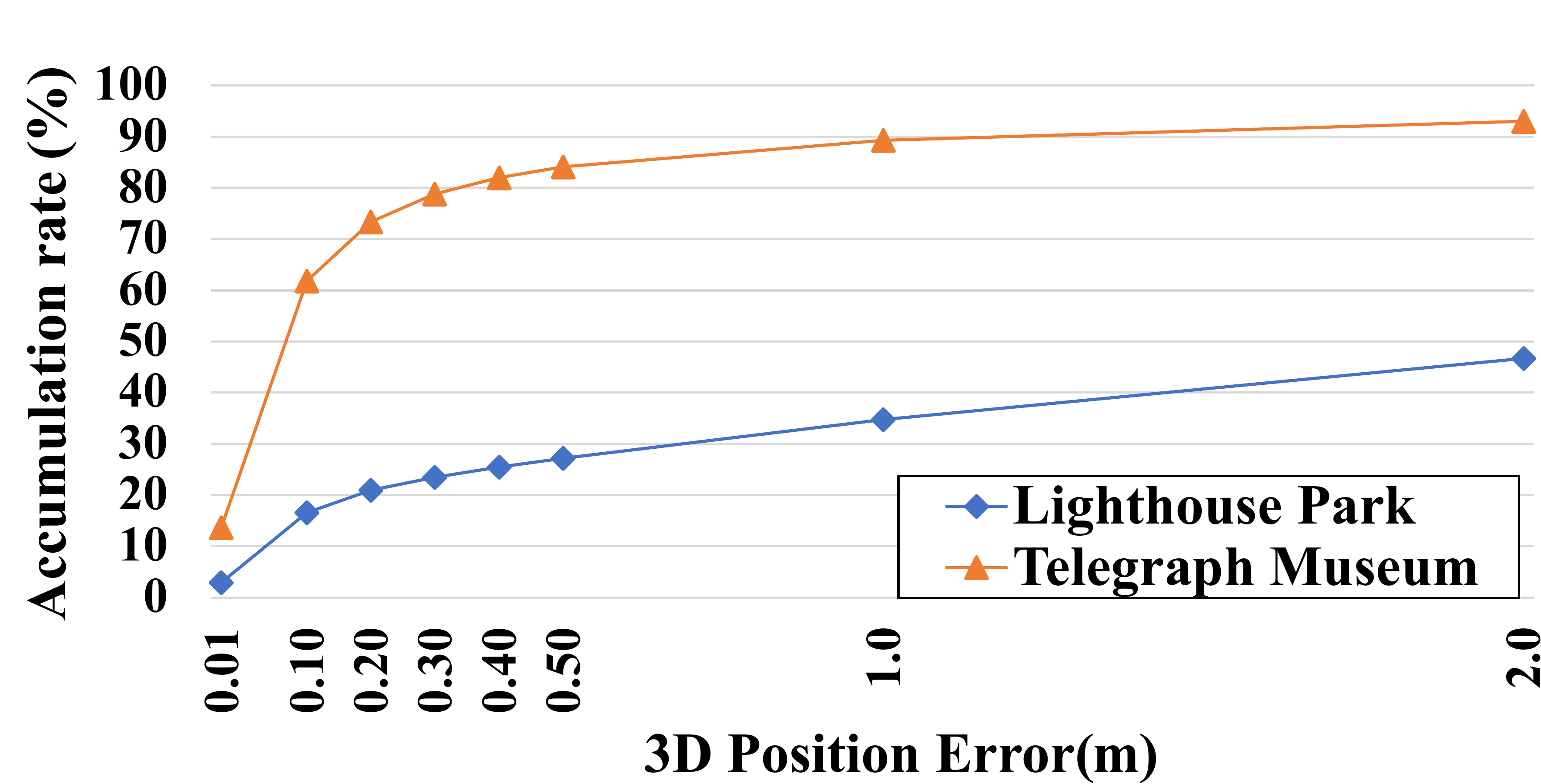}
  \caption{A graph showing the cumulative percentage of keypoints for which the 3D position error was below the threshold in the image-only method.}
  \label{fig:red80_shift}
\end{figure}

%% file: sec/5_Conclusion.tex
\section{Conclusion}
\label{sec:Conclusion}
In visual localization, we emphasize that the more keypoints in the 3D reference map and the lower the position error of the keypoints, the higher the accuracy of the camera pose estimation. We proposed LiM-Loc, which combines LiDAR point clouds and 2D keypoints to generate a dense and accurate 3D reference map. To directly assigned keypoints to the LiDAR point cloud with occlusions, we compress the point cloud into a spherical shell and remove invisible points. We evaluated LiM-Loc using indoor and outdoor datasets, and proved that a dense and accurate 3D reference map improves the accuracy of camera pose estimation. We applied our proposed method to several local features and showed that it improves their performance. We analyzed the results and confirmed the impact of the number and position errors of keypoints on the accuracy of camera pose transition. 

%% file: main.bbl
\begin{thebibliography}{39}
\providecommand{\natexlab}[1]{#1}
\providecommand{\url}[1]{\texttt{#1}}
\expandafter\ifx\csname urlstyle\endcsname\relax
  \providecommand{\doi}[1]{doi: #1}\else
  \providecommand{\doi}{doi: \begingroup \urlstyle{rm}\Url}\fi

\bibitem[Arandjelovic et~al.(2016)Arandjelovic, Gronat, Torii, Pajdla, and Sivic]{01}
Relja Arandjelovic, Petr Gronat, Akihiko Torii, Tomas Pajdla, and Josef Sivic.
\newblock Netvlad: Cnn architecture for weakly supervised place recognition.
\newblock In \emph{CVPR}, pages 5297--5307, 2016.

\bibitem[Berton et~al.(2023)Berton, Trivigno, Caputo, and Masone]{02}
Gabriele Berton, Gabriele Trivigno, Barbara Caputo, and Carlo Masone.
\newblock Training viewpoint robust models for visual place recognition.
\newblock In \emph{CVPR}, pages 11080--11090, 2023.

\bibitem[J. et~al.(2020)J., Pascal, and Eduard]{03}
Tyszkiewicz~Michal J., Fua Pascal, and Trulls Eduard.
\newblock Disk: learning local features with policy gradient.
\newblock In \emph{NeurIPS}, 2020.

\bibitem[Sarlin et~al.(2019)Sarlin, Cadena, Siegwart, and Dymczyk]{04}
Paul-Edouard Sarlin, Cesar Cadena, Roland Siegwart, and Marcin Dymczyk.
\newblock From coarse to fine: Robust hierarchical localization at large scale.
\newblock In \emph{CVPR}, pages 12716--12725, 2019.

\bibitem[Sarlin et~al.(2020)Sarlin, DeTone, Malisiewicz, and Rabinovich]{05}
Paul-Edouard Sarlin, Daniel DeTone, Tomasz Malisiewicz, and Andrew Rabinovich.
\newblock Superglue: Learning feature matching with graph neural networks.
\newblock In \emph{CVPR}, pages 12716--12725, 2020.

\bibitem[Zhao et~al.(2023)Zhao, Wu, Chen, Chen, Xu, and Li]{06}
Xiaoming Zhao, Xingming Wu, Weihai Chen, Peter C.~Y. Chen, Qingsong Xu, and Zhengguo Li.
\newblock Aliked: A lighter keypoint and descriptor extraction network via deformable transformation.
\newblock \emph{IEEE Transactions on Instrumentation and Measurement}, 72:\penalty0 1--16, 2023.

\bibitem[Hyeon et~al.(2021)Hyeon, Kim, and Doh]{07}
Janghun Hyeon, Joohyung Kim, and Nakju Doh.
\newblock Pose correction for highly accurate visual localization in largescale indoor spaces.
\newblock In \emph{ICCV}, pages 15974--15983, 2021.

\bibitem[Agarwal et~al.(2009)Agarwal, Furukawa, Snavely, Simon, Curless, Seitz, and Szeliski]{40}
Sameer Agarwal, Yasutaka Furukawa, Noah Snavely, Ian Simon, Brian Curless, Steven~M Seitz, and Richard Szeliski.
\newblock Building rome in a day.
\newblock In \emph{2009 IEEE 12th International Conference on Computer Vision}, pages 72--79, 2009.

\bibitem[Chen et~al.(2022)Chen, Wang, Wang, Tian, Xiong, and Li]{08}
Hansheng Chen, Pichao Wang, Fan Wang, Wei Tian, Lu Xiong, and Hao Li.
\newblock Epro-pnp: Generalized end-to-end probabilistic perspective-n-points for monocular object pose estimation.
\newblock In \emph{CVPR}, 2022.

\bibitem[Lepetit et~al.(2009)Lepetit, Moreno-Nogue, and Fua]{09}
Vincent Lepetit, Francesc Moreno-Nogue, and Pascal Fua.
\newblock Epnp: An accurate o(n) solution to the pnp problem.
\newblock \emph{Int. J. Comput. Vision}, 81\penalty0 (2):\penalty0 155--166, 2009.

\bibitem[Schönberger and Frahm(2016)]{10}
Johannes~L. Schönberger and Jan-Michael Frahm.
\newblock Structure-from-motion revisited.
\newblock In \emph{CVPR}, pages 4104--4113, 2016.

\bibitem[Zhang et~al.(2022)Zhang, Yang, Zhang, and Zhang]{13}
X. Zhang, J. Yang, S. Zhang, and Y. Zhang.
\newblock 3d registration with maximal cliques.
\newblock In \emph{CVPR}, pages 17745--17754, 2022.

\bibitem[Besl and McKay(1992)]{14}
Paul~J. Besl and Neil~D. McKay.
\newblock A method for registration of 3d shapes.
\newblock \emph{IEEE Transactions on Pattern Analysis and Machine Intelligence}, 14\penalty0 (2):\penalty0 239--256, 1992.

\bibitem[Fischler and Bolles(1981)]{12}
M.~A. Fischler and R.~C. Bolles.
\newblock Random sample consensus: a paradigm for model fitting with applications to image analysis and automated cartography.
\newblock \emph{Communications of the ACM}, 24\penalty0 (6):\penalty0 381--395, 1981.

\bibitem[Ge et~al.(2020)Ge, Wang, Zhu, Zhao, and Li]{15}
Yixiao Ge, Haibo Wang, Feng Zhu, Rui Zhao, and Hongsheng Li.
\newblock Self-supervising fine-grained region similarities for large-scale image localization.
\newblock In \emph{ECCV}, 2020.

\bibitem[Revaud(2019)]{16}
Jerome Revaud.
\newblock Learning with average precision: Training image retrieval with a listwise loss.
\newblock In \emph{ICCV}, pages 5106--5115, 2019.

\bibitem[Berton et~al.(2022)Berton, Masone, and Caputo]{17}
Gabriele Berton, Carlo Masone, and Barbara Caputo.
\newblock Rethinking visual geo-localization for large-scale applications.
\newblock In \emph{CVPR}, pages 4878--4888, 2022.

\bibitem[Dusmanu et~al.(2019)Dusmanu, Rocco, Pajdla, Pollefeys, Sivic, Torii, and Sattler]{19}
Mihai Dusmanu, Ignacio Rocco, Tomas Pajdla, Marc Pollefeys, Josef Sivic, Akihiko Torii, and Torsten Sattler.
\newblock D2-net: A trainable cnn for joint detection and description of loca features.
\newblock In \emph{CVPR}, 2019.

\bibitem[Revaud et~al.(2019)Revaud, Weinzaepfel, de~Souza, and Humenberger]{20}
Jerome Revaud, Philippe Weinzaepfel, C'esar~Roberto de Souza, and Martin Humenberger.
\newblock R2d2: repeatable and reliable detector and descriptor.
\newblock In \emph{NeurIPS}, 2019.

\bibitem[Maddern et~al.(2017)Maddern, Pascoe, Linegar, and Newman]{29}
Will Maddern, Geoffrey Pascoe, Chris Linegar, and Paul Newman.
\newblock 1 year, 1000 km: The oxford robotcar dataset.
\newblock \emph{Int. J. Rob. Res.}, 36\penalty0 (1):\penalty0 3--15, 2017.

\bibitem[DeTone et~al.(2018)DeTone, Malisiewicz, and Rabinovich]{31}
Daniel DeTone, Tomasz Malisiewicz, and Andrew Rabinovich.
\newblock Superpoint: Self-supervised interest point detection and description.
\newblock In \emph{CVPR}, pages 224--236, 2018.

\bibitem[Lowe(2004)]{18}
David~G Lowe.
\newblock Distinctive image features from scaleinvariant keypoints.
\newblock \emph{International Journal of Computer Vision}, 60:\penalty0 91--110, 2004.

\bibitem[Rocco et~al.(2018)Rocco, Cimpoi, Arandjelovic, Torii, Pajdla, and Sivic]{33}
I. Rocco, M. Cimpoi, R. Arandjelovic, A. Torii, T. Pajdla, and J. Sivic.
\newblock Neighbourhood consensus networks.
\newblock In \emph{NeurIPS}, 2018.

\bibitem[Rocco et~al.(2020)Rocco, Arandjelovic, and Sivic]{34}
Ignacio Rocco, Relja Arandjelovic, and Josef Sivic.
\newblock Efficient neighbourhood consensus networks via submanifold sparse convolutions.
\newblock In \emph{ECCV}, pages 605--621, 2020.

\bibitem[Tinchev et~al.(2021)Tinchev, Li, Han, Mitchell, and Kouskouridas]{35}
Georgi Tinchev, Shuda Li, Kai Han, David Mitchell, and Rigas Kouskouridas.
\newblock Xresolution correspondence networks.
\newblock In \emph{British Machine Vision Conference (BMVC)}, 2021.

\bibitem[Zhou et~al.(2021)Zhou, Sattler, and Leal-Taixe]{36}
Qunjie Zhou, Torsten Sattler, and Laura Leal-Taixe.
\newblock Patch2pix: Epipolar-guided pixel-level correspondences.
\newblock In \emph{CVPR}, pages 4669--4678, 2021.

\bibitem[Kuang et~al.(2022)Kuang, Li, He, Wang, and Zhao]{39}
Zhengfei Kuang, Jiaman Li, Mingming He, Tong Wang, and Yajie Zhao.
\newblock Densegap: Graph-structured dense correspondence learning with anchor points.
\newblock In \emph{ICPR}, 2022.

\bibitem[Shen et~al.(2020)Shen, Darmon, Efros, and Aubry]{37}
Xi Shen, Francois Darmon, Alexei~A Efros, and Mathieu Aubry.
\newblock Ransac-flow: generic two-stage image alignment.
\newblock In \emph{ECCV}, pages 618--637, 2020.

\bibitem[Truong et~al.(2020)Truong, Danelljan, and Timofte]{38}
Prune Truong, Martin Danelljan, and Radu Timofte.
\newblock Glunet: Global-local universal network for dense flow and correspondences.
\newblock In \emph{CVPR}, pages 6258--6268, 2020.

\bibitem[Lindenberger et~al.(2023)Lindenberger, Sarlin, and Pollefeys]{22}
Philipp Lindenberger, Paul-Edouard Sarlin, and Marc Pollefeys.
\newblock Lightglue: Local feature matching at light speed.
\newblock In \emph{ICCV}, 2023.

\bibitem[Sun et~al.(2021)Sun, Shen, Wang, Bao, and Zhou]{21}
Jiaming Sun, Zehong Shen, Yuang Wang, Hujun Bao, and Xiaowei Zhou.
\newblock Loftr: Detector-free local feature matching with transformers.
\newblock In \emph{CVPR}, pages 8922--8931, 2021.

\bibitem[Bach et~al.(2022)Bach, Dinh, and Lee]{28}
Thuan~Bui Bach, Tuan~Tran Dinh, and Joo-Ho Lee.
\newblock Featloc: Absolute pose regressor for indoor 2d sparse features with simplistic view synthesizing.
\newblock \emph{ISPRS Journal of Photogrammetry and Remote Sensing}, 189:\penalty0 50--62, 2022.

\bibitem[Sattler et~al.(2019)Sattler, Zhou, Pollefeys, and Leal-Taixe]{41}
Torsten Sattler, Qunjie Zhou, Marc Pollefeys, and Laura Leal-Taixe.
\newblock Understanding the limitations of cnn-based absolute camera pose regression.
\newblock In \emph{CVPR}, pages 3302--3312, 2019.

\bibitem[Kendall and Cipolla(2017)]{42}
Alex Kendall and Roberto Cipolla.
\newblock Geometric loss functions for camera pose regression with deep learning.
\newblock In \emph{CVPR}, page 5974–5983, 2017.

\bibitem[Katz et~al.(2007)Katz, Tal, and Basri]{23}
Sagi Katz, Ayellet Tal, and Ronen Basri.
\newblock Direct visibility of point sets.
\newblock In \emph{SIGGRAPH}, 2007.

\bibitem[Geiger et~al.(2012)Geiger, Lenz, and Urtasun]{24}
Andreas Geiger, Philip Lenz, and Raquel Urtasun.
\newblock Are we ready for autonomous driving? the kitti vision benchmark suite.
\newblock In \emph{CVPR}, 2012.

\bibitem[Liao et~al.(2023)Liao, Xie, and Geiger]{25}
Yiyi Liao, Jun Xie, and Andreas Geiger.
\newblock Kitti-360: A novel dataset and benchmarks for urban scene understanding in 2d and 3d.
\newblock \emph{IEEE Transactions on Pattern Analysis and Machine Intelligence}, 45\penalty0 (3):\penalty0 3292--3310, 2023.

\bibitem[Taira et~al.(2018)Taira, Okutomi, Sattler, Cimpoi, Pollefeys, Sivic, Pajdla, and Torii]{26}
Hajime Taira, Masatoshi Okutomi, Torsten Sattler, Mircea Cimpoi, Marc Pollefeys, Josef Sivic, Tomas Pajdla, and Akihiko Torii.
\newblock Inloc: Indoor visual localization with dense matching and view synthesis.
\newblock In \emph{CVPR}, 2018.

\bibitem[Armeni et~al.(2017)Armeni, Sax, Zamir, and Savarese]{27}
I. Armeni, A. Sax, A.~R. Zamir, and S. Savarese.
\newblock Joint 2d- 3d-semantic data for indoor scene understanding, 2017.
\newblock ArXiv e-prints.

\end{thebibliography}
